\documentclass{article}
\usepackage{spconf,amsmath,graphicx}

\title{Improving Catheter Segmentation \& Localization\\ in 3D Cardiac Ultrasound Using Direction-Fused FCN}
\name{Hongxu Yang$^{*}$, Caifeng Shan$^{\dagger}$, Alexander F. Kolen$^{\dagger}$, Peter H.N. de With$^{*}$}
\address{$^{*}$Eindhoven University of Technology, Eindhoven, The Netherlands\\
$^{\dagger}$Philips Research, Eindhoven, The Netherlands}
\begin{document}
%
\maketitle
\begin{abstract}
Fast and accurate catheter detection in cardiac catheterization using harmless 3D ultrasound (US) can improve the efficiency and outcome of the intervention. However, the low image quality of US requires extra training for sonographers to localize the catheter. In this paper, we propose a catheter detection method based on a pre-trained VGG network, which exploits 3D information through re-organized cross-sections to segment the catheter by a shared fully convolutional network (FCN), which is called a Direction-Fused FCN (DF-FCN). Based on the segmented image of DF-FCN, the catheter can be localized by model fitting. Our experiments show that the proposed method can successfully detect an ablation catheter in a challenging \textit{ex-vivo} 3D US dataset, which was collected on the porcine heart. Extensive analysis shows that the proposed method achieves a Dice score of 57.7\%, which offers at least an 11.8\% improvement when compared to state-of-the-art instrument detection methods. Due to the improved segmentation performance by the DF-FCN, the catheter can be localized with an error of only 1.4~mm.
\end{abstract}
\begin{keywords}
3D ultrasound, catheter segmentation \& localization, VGG pre-trained model, fine-tuning.
\end{keywords}
\section{Introduction}
Cardiac catheterization has been extensively applied during interventional therapy for heart diseases, which offers a short recovery period and low risk for the patients. Because a direct view on the organ is occluded by skin or other tissues during the operation, medical imaging methods, such as fluoroscopic (X-ray) or ultrasound (US) imaging, are used to guide the intervention. The maturity of 3D cardiac US imaging, e.g. Transesophageal Echography (TEE), makes 3D US-guided operations an attractive option because of its lower radiation exposure for both patient and surgeon. Besides this, 3D US also provides richer spatial information of the tissue than the traditional 2D X-ray imaging. Unfortunately, low contrast 3D US with anatomical structure makes it difficult for the physician or sonographer to timely localize the catheter during intervention.  Therefore, computer-aided catheter detection (CAD) in 3D US can be helpful for the physician and make the operation more efficient.

Techniques like active sensing \cite{sensor} or robotics \cite{robotic}, have introduced extra cost for equipment and training. Alternatively, direct image-based catheter detection will lead to a lower cost for hospitals. Catheter detection in US has therefore been studied in literature. A template matching approach was proposed by Cao \cite{automated}. Alternatively, the Frangi feature was introduced with supervised learning methods and shown to give successful detection for instruments \cite{uhervcik2013line}.  A recent study on instrument detection combined Gabor features with Frangi features for catheter detection in phantom heart data, which showed promising results in Phantom~\cite{pourtaherian2017medical}. More recently, more discriminative features were introduced by \cite{OURs} and showed a clear improvement in catheter detection using \textit{ex-vivo} data. 

\begin{figure*}[ht]
\begin{minipage}[b]{1.0\linewidth}
  \centering
  \centerline{\includegraphics[width=15cm]{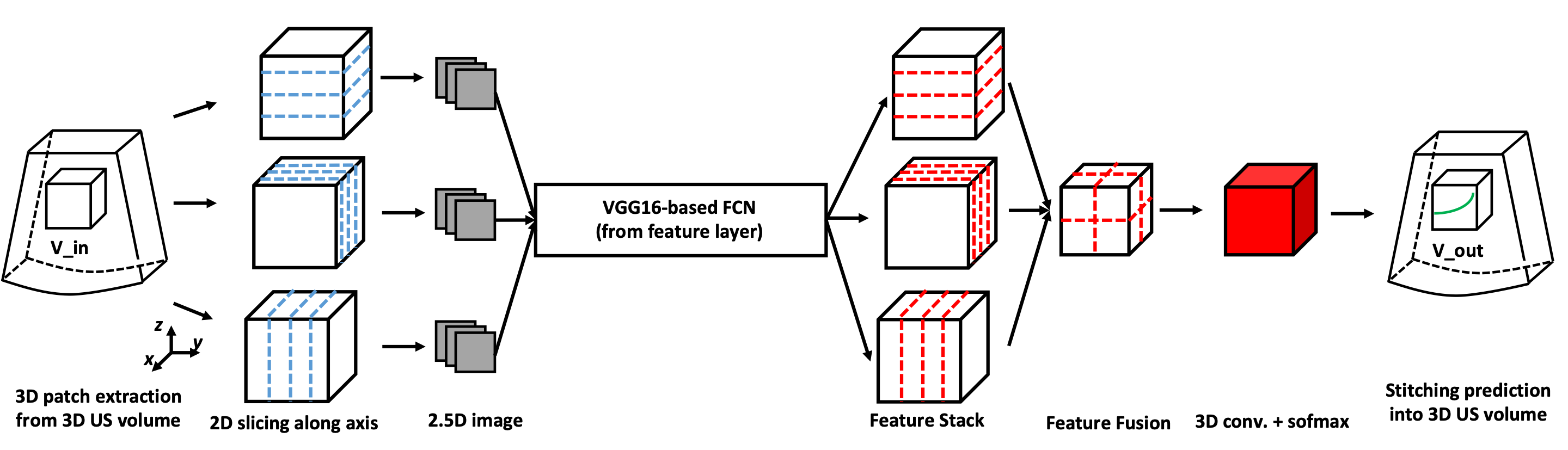}}
  \centerline{(a) Patch-based catheter segmentation in 3D US}\medskip
\end{minipage}
\begin{minipage}[b]{1.0\linewidth}
  \centering
  \centerline{\includegraphics[width=15cm]{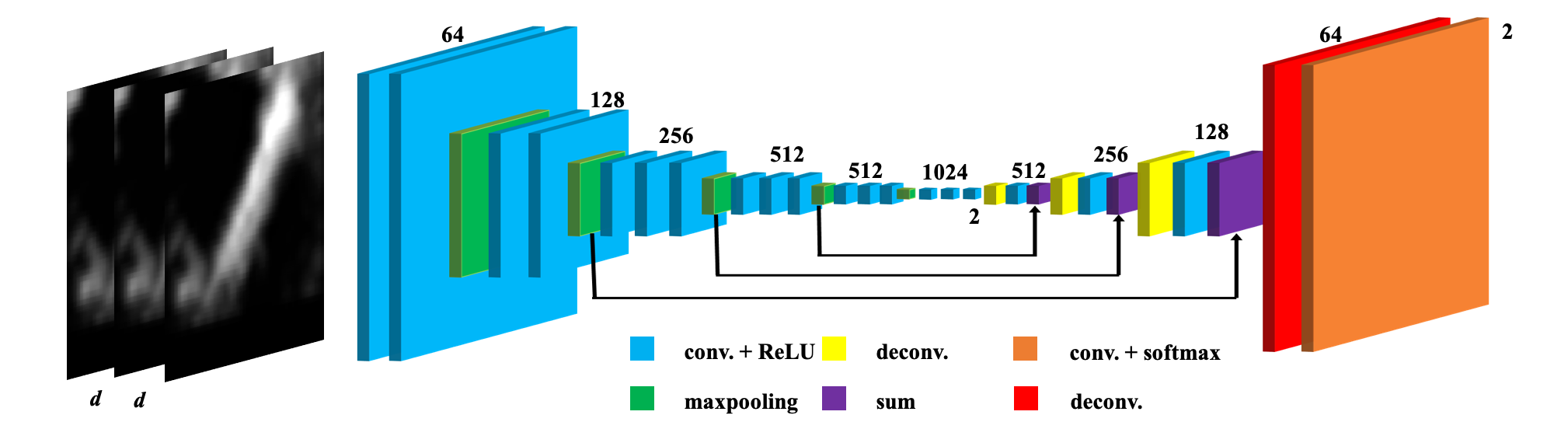}}
  \centerline{(b) VGG16-based FCN in 2D image}\medskip
\end{minipage}
\caption{Overview of patch-based catheter segmentation in 3D US. (a) Patch-based catheter segmentation using Direction-Fused FCN. (b) Structure of the FCN segmentation based on the VGG network for RGB images, where the $d$ represents the voxel gap between slices and the red block at deconv. indicates the output feature layer for feature fusion. }
\label{fig:overview}
\end{figure*}

In recent years, deep learning methods, like convolutional neural networks (CNN), have shown a performance improvement in medical instruments detection, such as catheter/needle detection in 3D US~\cite{pourtaherian2017improving}\cite{TriHang2018}. Nevertheless, the local voxel-wise processing in \cite{pourtaherian2017improving}\cite{TriHang2018} requires the CNN to label the 3D image in a voxel-by-voxel manner, which gives 3D contextual information loss and more computation time. Additionally, the continuous work of Poutaherian \textit{et al.}~\cite{pourtaherian2018robust} demonstrates a needle segmentation using a shared 2D fully convolutional network, ShareFCN \cite{pourtaherian2018robust}\cite{FCN}, together with (image) plane-based segmentation. However, this approach with plane-based processing cannot fully exploit the 3D information by simply fusing the probability estimation through averaging along two axes. To better exploit the 3D information while preserving the advantages of the 2D pre-trained model, we propose a catheter segmentation method in 3D US by extracting its learned 2D features along the principal axes, based on which 2D features are fused to boost the performance. Meanwhile, we introduce the U-Net structure into the FCN network to exploit intra-plane information ranging from local textures up to semantic contextual information~\cite{UNET}. With extensive validation on the challenging \textit{ex-vivo} dataset, our system obtains a promising Dice score at 57.7\%, which offers at least an 11.8\% improvement when compared to \cite{OURs}\cite{pourtaherian2017improving}\cite{TriHang2018}. Based on the segmented catheter in 3D US, its skeleton can be correctly localized with an average error of only 1.4 mm, which is less than catheter diameter.
\section{Methods}
Fig.~\ref{fig:overview} illustrates the framework for catheter segmentation in 3D US. The full volume US is divided into smaller patches $V_{in}$. The patch is first sliced into neighborhood pseudo-RGB images by using a spatial gap $d$ in each direction. Then, the images are processed as in Fig.~\ref{fig:overview}\,(b), which is an FCN with U-Net like structure, partly based on a pre-trained VGG-16 network~\cite{VGG16}. Based on the prediction from stacked features, the predicted patches are stitched back into the 3D volume to provide a final result of the catheter segmentation. More details of the semantic segmentation in the 3D patch and VGG16-based FCN are discussed below.
\subsection{Catheter segmentation using Direction-Fused FCN}
The 3D US volume is divided into several smaller patches, i.e. $V_{in}$, where each $V_{in}$ is sliced into several 2D (RGB) images along each axis. For example, a patch $V_{in}$ with size $48\times48\times48$ is decomposed into 48 planes along each axis. For each plane, a 3-channel image is constructed based on the adjacent images of the actual plane with spatial gap $d$ voxels along the axis, which is shown in Fig.~\ref{fig:overview} (b). As a result, the patch $V_{in}$ leads to the generation of 48 different 3-channel images in each direction (padding is applied at the boundary plane). Then, each image is processed by the 2D FCN. As shown in Fig.~\ref{fig:overview} (a), the images are processed by the FCN and its output features for each image (from the red block) are stacked based on the plane's original position to construct feature maps in three axes (a high dimensional transpose operation is applied). Further, feature maps from different axes are accumulated to form a fused feature map. Finally, the final prediction is obtained by applying a 3D convolution and a softmax layer. This approach can exploit 3D information in both inter-slice and intra-slice by the 2.5D manner. Because this structure fused the 2D feature maps into 2.5D maps through three different directions, we denote it as Direction-Fused FCN (DF-FCN). As a comparison to show the ability of DF-FCN, we also consider a simpler case as baseline, just FCN. Then, the FCN predicts each 2D image directly along the axis and stacks the 2D predictions back to the 3D patch (the predictions are applied along one of the three axes, and randomly chosen). This approach only makes use of 3D information by introducing the voxel gap~$d$.
\subsection{VGG16-based FCN and training}
Fig.~\ref{fig:overview}\,(b) shows the VGG16-based FCN architecture of our work, which consists of a convolution and a deconvolution part. The convolution layers with 5 maxpooling layers are the same as with VGG-16 \cite{VGG16}. Then, the output of the last maxpooling layer is filtered by 3 convolutional layers with filter size 1024, 1024 and 2. The deconvolution part includes 4 deconvolution layers with masks of $2\times2$, $2\times2$, $2\times2$ and $4\times4$ pixels. Furthermore, after each deconvolution layer, an extra convolution operation is included to improve the performance \cite{UNET}. To reduce the usage of the GPU memory, we employ summation instead of concatenation in the deconvolution part. The output of the VGG16-based FCN is the probability prediction after the softmax layer or 2D features from ReLU layer~\cite{ReLU}, which is indicated by the red square in Fig.~\ref{fig:overview}\,(b).

During the training, the parameters of the convolution part in the FCN are initialized using the pre-trained VGG16 model. The remaining parameters in the deconvolution part are randomly initialized. In order to generalize the network, data augmentations like rotation, mirroring, contrast and brightness transformations are performed randomly on-the-fly on each input patch. To augment the training images, the training patches are extracted based on annotated catheter voxels in 3D US, i.e. a patch is extracted from 3D US based on an annotated catheter voxel using it as the patch center. Meanwhile, non-catheter voxels are randomly downsampled to generate the same amount of non-catheter patches, which may not include the catheter or only partly include the catheter. All parameters of the FCN are fine-tuned or learned by minimizing the cross-entropy using the Adam optimizer~\cite{Adam} with an initial learning rate of $10^{-5}$, based on the 3D patch $V_{in}$ and its prediction. Furthermore, to avoid over-fitting, dropout layers~\cite{DO} are introduced with an empirical probability 0.85 after convolution layer 14 and 15, see \cite{VGG16}. 
\subsection{Catheter localization by model fitting}
The segmented volume includes some false positives due to complex anatomical structures in the data volume. To accurately localize the catheter, we adapt the catheter model-fitting algorithm according to the Sparse-Plus-Dense (SPD) RANSAC algorithm~\cite{TriHang2018}\cite{SPD}. In the segmented binary image, named dense volume, the voxels are clustered through connectivity analysis. The skeleton of each cluster is extracted along the dominant direction of that cluster. These skeletons generate a sparse volume. When fitting the catheter model, three control points are randomly selected and re-ranked by direction analysis from the sparse volume. Then a cubic spline is applied to generate the catheter model. The model of the dense volume that includes the highest number of dense voxels is chosen as the catheter. In this paper, the threshold controlling the ratio of inliers and outliers is chosen as three voxels. The SPD algorithm avoids redundancy in the selection of boundary points during RANSAC iterations \cite{SPD}.

\subsection{Testing and evaluation metrics}
During the prediction, the observed volume is divided into adjacent patches with size $N\times{N}\times{N}$ voxels. Then, these patches are enlarged by including their neighborhood voxels around them in the original volume, which finally leads to size $M\times{M}\times{M}$ voxels with $N\le{M}$. Based on the prediction of patch $V_{out}$, its center volume with size $N\times{N}\times{N}$ is extracted for 3D US reconstruction. 

To evaluate the segmentation performance on the whole volume with imbalanced classes distribution, the metrics of Recall, Precision, Dice coefficient (Dice), and Average Hausdorff Distance (AHD) are employed~\cite{taha2015metrics}. The catheter localization accuracy is measured by the skeleton error (SE) and the position errors of the two endpoints (which is averaged as 'endpoints' error, termed EE). The SE is the average distance between 5 equally-sampled points on the fitted skeleton and the annotation skeleton. The EE is the average distance between the endpoints on the detected catheter and the corresponding points on the annotation.

\section{Experiments}
\begin{table*}[ht]
\centering
\caption{Average performance of catheter segmentation \& localization (mean$\pm$ std. ).}
\label{SegPer}
\begin{tabular}{c|cccc|cc}
\hline
Method & Recall (\%) & Precision (\%) & Dice (\%)& AHD (voxel)&SE (mm)&EE (mm)\\ \hline
Handcrafted~\cite{OURs}&45.5$\pm$26.3&23.2$\pm$11.5&29.4$\pm$15.3&10.2$\pm$6.8&4.0$\pm$4.4&4.6$\pm$4.3\\ \hline
ShareConv~\cite{pourtaherian2017improving}&41.4$\pm$14.6&55.9$\pm$10.9&45.9$\pm$11.3&4.3$\pm$3.3&1.6$\pm$0.8&2.2$\pm$0.8\\ \hline
LateFusion~\cite{TriHang2018}&42.8$\pm$14.1&70.1$\pm$12.2&51.6$\pm$11.4&1.6$\pm$8.7&1.6$\pm$1.2&2.3$\pm$1.2\\ \hline
DF-FCN&50.1$\pm$17.2&72.5$\pm$11.6&57.7$\pm$12.0&1.0$\pm$0.5&1.4$\pm$0.5&1.8$\pm$1.1\\ \hline
\end{tabular}
\end{table*}

The proposed method was evaluated on a 3D US dataset of isolated porcine heart (approved by review board), which was collected by a 2-7 MHz phase array transducer (TEE) when inserting an ablation catheter (Boston Scientific, diameter 2.3 mm) into the left ventricle or right ventricle. The dataset includes 25 images, individually extracted from 25 separate 3-second videos. The original voxels were re-sampled to obtain an isotropic voxel size in each direction (volume size: $128\times128\times128$ voxels, space unit: 0.54 mm/direction). The volumes were manually annotated and partitioned into two classes: catheter and non-catheter (confirmed by experts). During the experiment, we empirically used a 3-fold cross-validation to separate training and testing data. Due to the spatial resolution and trade-off of the GPU memory, we have empirically chosen $M=48$ and $ N=32$ in our experiment to fit the GPU memory. Meanwhile, to fully exploit the influence of voxel gap $d$, we have also performed validation on it with different integer values from 0 to~5. All the experiments were performed on a GeForce GTX 1080Ti.

\subsection{Spatial representation of fine-tuned DF-FCN}
\begin{figure}[ht]
\centering{\includegraphics[width=8cm]{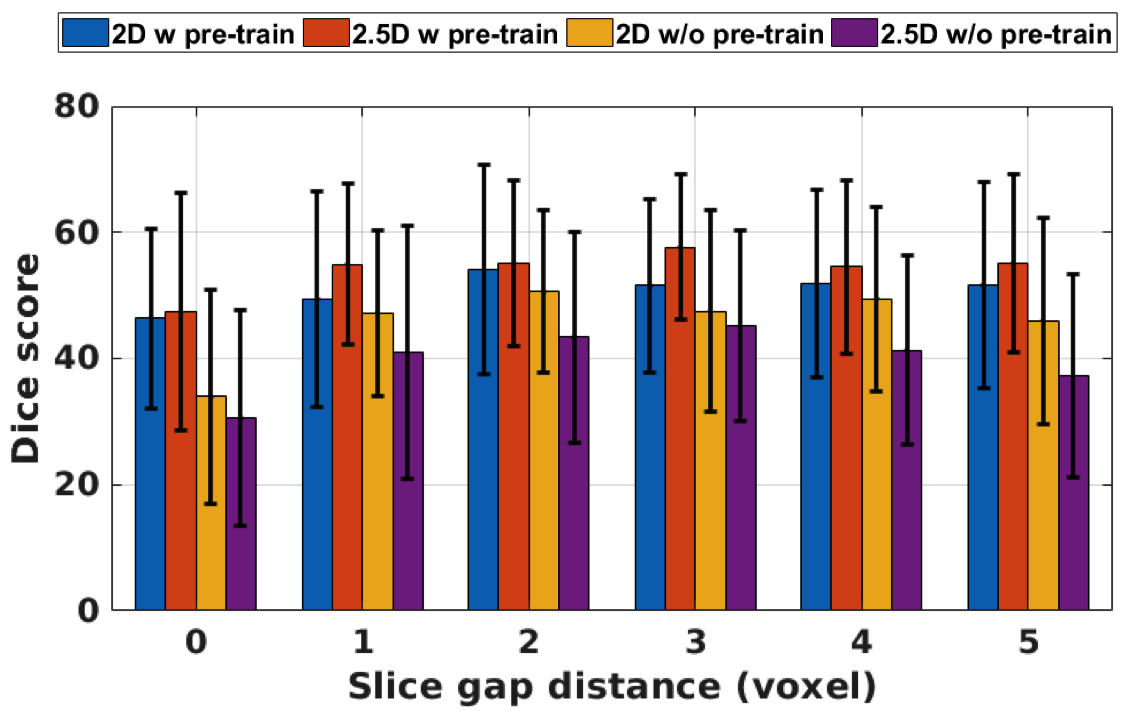}}
\caption{Segmentation performance with respect to voxel gap distance in two approaches.}
\label{fig:Spatial}
\vspace*{-0.2cm}
\end{figure}
First, our proposed systems are compared based on different values of $d$ in FCN and DF-FCN. The methods using 2D FCN prediction with/without pre-training is abbreviated to "2D w(/o) pre-train", while the DF-FCN using 2D feature fusion is denoted by "2.5D w(/o) pre-train". Their performances with respect to the voxel gap $d$ are depicted in Fig.~\ref{fig:Spatial} using the Dice score. From the figure, several conclusions can be made. First, with a pre-trained model, the network achieves better performance. Meanwhile, the 2.5D approach without pre-training has the worst performance due to the training of a much more complex network, when compared to the 2D without pre-training approach. Second, with increasing $d$, the performances are first increasing, but finally degrading with a large distance. This is caused by a too large gap which make the catheter invisible for the network, so that the spatial correlation is decreased. Third, the performance based on 2.5D features has a better and more robust performance when compared with a single direction method. This is evident and due to the increased use of 3D information, which re-organizes the segmentation of the catheter in 3D US. From all cases, the 2.5D with pre-trained model performs the highest Dice when~$d=3$. 

\subsection{Segmentation and localization performance}
Based on DF-FCN with a pre-trained model, our proposed method is compared to a method using multi-scale handcrafted features with an AdaBoost classifier~\cite{OURs} and methods for medical instrument segmentation based on voxel-wise tri-planar CNN classification (ShareConv and LateFusion)~\cite{pourtaherian2017improving}\cite{TriHang2018}. With respect to the handcrafted system \cite{OURs}, the performances were obtained by tuning the threshold to achieve the highest Dice score in each fold. Besides, we do not consider the weighted loss function from~\cite{TriHang2018} to show fairness in voxel distribution. The catheter segmentation performances are shown in Table~\ref{SegPer}. Furthermore, the catheter localization performances are shown in the table expressed in millimeters/voxels using the earlier proposed localization metrics. 

\begin{figure}[ht]
\centering{\includegraphics[width=7cm]{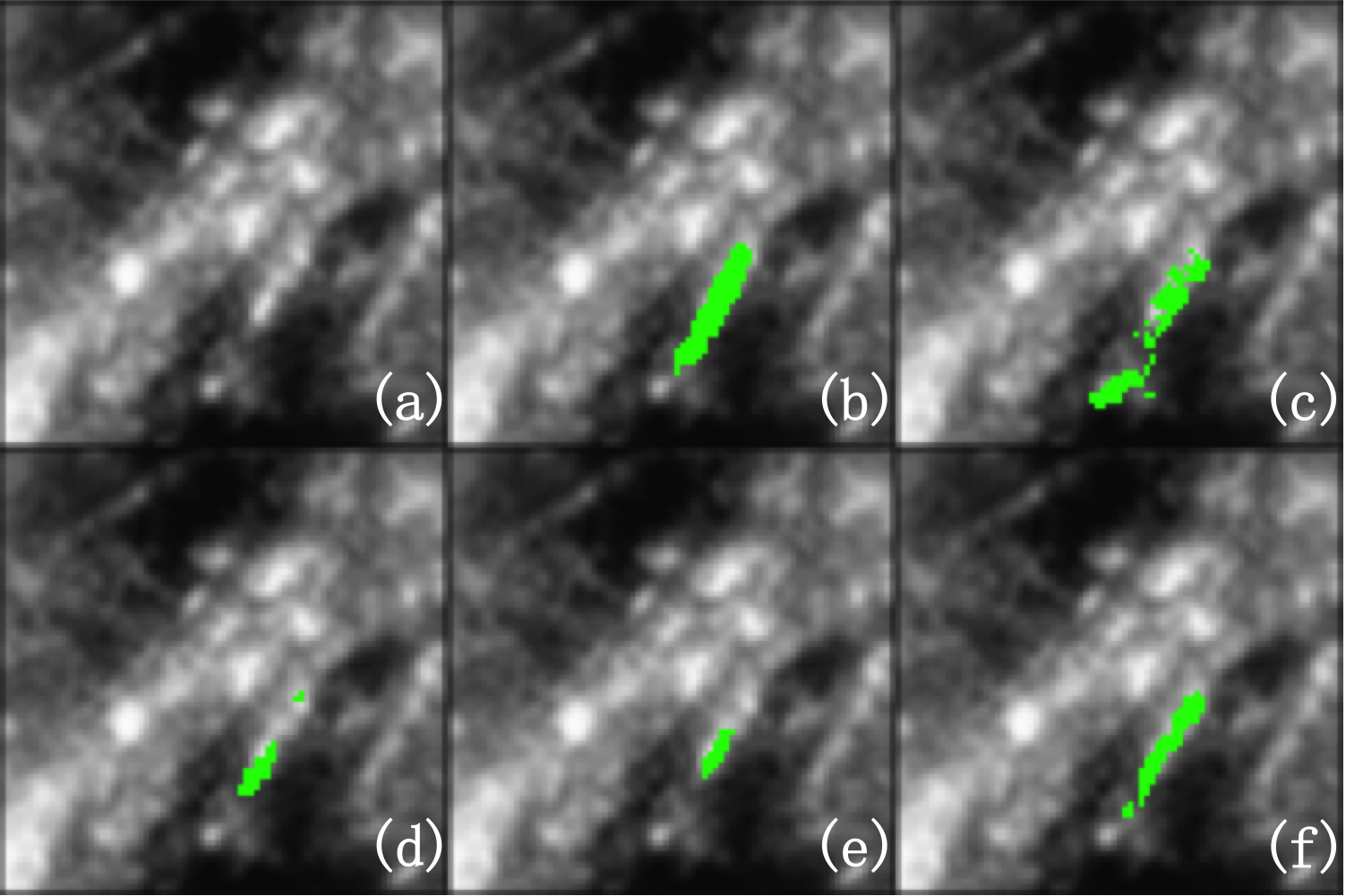}}
\caption{Catheter segmentation under different approaches. (a) original image, (b) annotation, (c) handcrafted method, (d) LateFusion CNN, (e) FCN and (f) DF-FCN.} 
\label{fig:segment}
\end{figure}

Considering Table~\ref{SegPer}, DF-FCN achieves the best performance when compared with handcrafted features and voxel-wise CNNs with respect to all segmentation metrics (p-value$<$0.05). Our proposed method is able to achieve better catheter localization results, which rely on a better segmentation using more contextual information. Furthermore, due to semantic segmentation, the patch-wise segmentation requires less inference time when compared to the voxel-level classification (from several min. to less than one min.). Fig.~\ref{fig:segment} depicts some segmentation results. From the comparison, it is found that with the limited contextual information in LateFusion CNN and FCN, these methods cannot recognize the boundary voxels of the catheter in challenging US images. Although the handcrafted method is able to classify more voxels as part of the catheter, too much false positives lead to more difficulties in catheter localization.
\section{Conclusions}
We have proposed a catheter detection method for 3D US images with improved detection accuracy. This method is based on learning a fine-tuned FCN with feature maps fused along principal directions, which is combined with catheter segmentation and model-based localization. The proposed method was evaluated on an \textit{ex-vivo} dataset, which showed significant improvement when compared to the state-of-the-art methods (achieved Dice score of 57.7\% with 11.8\% improvement). The results show that our method can localize a catheter with an error less than the catheter diameter in a complex anatomical 3D US image. Therefore, the proposed catheter detection method can facilitate surgeons to interpret the catheter during interventions, which reduces the operation time and provides a better outcome for the patient. 
\bibliographystyle{IEEEbib}
\bibliographystyle{unsrt}

\end{document}